\documentclass[runningheads]{llncs}
\usepackage[T1]{fontenc}

\usepackage{graphicx}
\usepackage{hyperref}
\usepackage{url}
\usepackage{booktabs}
\usepackage[autolanguage]{numprint}
\usepackage{siunitx}
\usepackage{multirow}
\usepackage{microtype}
\usepackage[dvipsnames]{xcolor}
\usepackage{amssymb}

\usepackage{amsmath}
\usepackage{amsfonts}

\usepackage{ifthen}
\newboolean{authorv}
\setboolean{authorv}{true}

\ifthenelse{\boolean{authorv}}{
\hypersetup{
    colorlinks=true,
    linkcolor=[RGB]{0 0 0},
    urlcolor=[RGB]{0 0 0},
    citecolor=[RGB]{0 0 0},
    pdfborder={0 0 0},
    frenchlinks=false
}
\usepackage{orcidlink}
\renewcommand{\orcidID}{\orcidlink}
}

\newcommand{\clipv}{\mathrm{CLIP}_V}
\newcommand{\clipt}{\mathrm{CLIP}_T}

\newcommand{\tqtp}{TqTp}
\newcommand{\tqiq}{TqIq}
\newcommand{\iqip}{IqIp}
\newcommand{\tpip}{TpIp}
\newcommand{\iqtp}{IqTp}

\newcommand{\TQTP}{\textsc{\tqtp{}}}
\newcommand{\TQIQ}{\textsc{\tqiq{}}}
\newcommand{\IQIP}{\textsc{\iqip{}}}
\newcommand{\TPIP}{\textsc{\tpip{}}}
\newcommand{\IQTP}{\textsc{\iqtp{}}}

\begin{document}
\title{Cross-modal Retrieval for \\Knowledge-based Visual Question Answering\thanks{
We thank the anonymous reviewers for their helpful comments, as well as Antoine Chaffin for fruitful discussions about CLIP and cross-modal retrieval. 
Paul Lerner did this work during his PhD at LISN. 
This work was supported by the ANR-19-CE23-0028 MEERQAT project. This work was granted access to the HPC resources of IDRIS under the allocation 2021-AD011012846 made by GENCI. 
}
}
\author{Paul Lerner\inst{1}\orcidID{0000-0002-0882-8684} \and
Olivier Ferret\inst{2}\orcidID{0000-0003-0755-2361} \and
Camille Guinaudeau\inst{3}\orcidID{0000-0001-7249-8715}}
\authorrunning{P. Lerner et al.}%
\institute{
Sorbonne Université, CNRS, ISIR, 75005, Paris, France \\
\email{lerner@isir.upmc.fr}
\and
Université Paris-Saclay, CEA, List, F-91120, Palaiseau, France\\
\email{olivier.ferret@cea.fr}
\and
Université Paris-Saclay, CNRS, LISN, 91400, Orsay, France \\
\email{camille.guinaudeau@lisn.upsaclay.fr}
}
\maketitle              %
\begin{abstract}
Knowledge-based Visual Question Answering about Named Entities is a challenging task that requires retrieving information from a multimodal Knowledge Base. 
Named entities have diverse visual representations and are therefore difficult to recognize. We argue that cross-modal retrieval may help bridge the semantic gap between an entity and its depictions, and is foremost complementary with mono-modal retrieval. We provide empirical evidence through experiments with a multimodal dual encoder, namely  CLIP, on the recent ViQuAE,  InfoSeek, and Encyclopedic-VQA datasets. 
Additionally, we study three different strategies to fine-tune such a model: mono-modal, cross-modal, or joint training. 
Our method, which combines mono- and cross-modal retrieval, is competitive with billion-parameter models on the three datasets, while being conceptually simpler and computationally cheaper. 

\end{abstract}

\section{Introduction}
\begin{figure}[t]
\centering
\includegraphics[width=.9\textwidth]{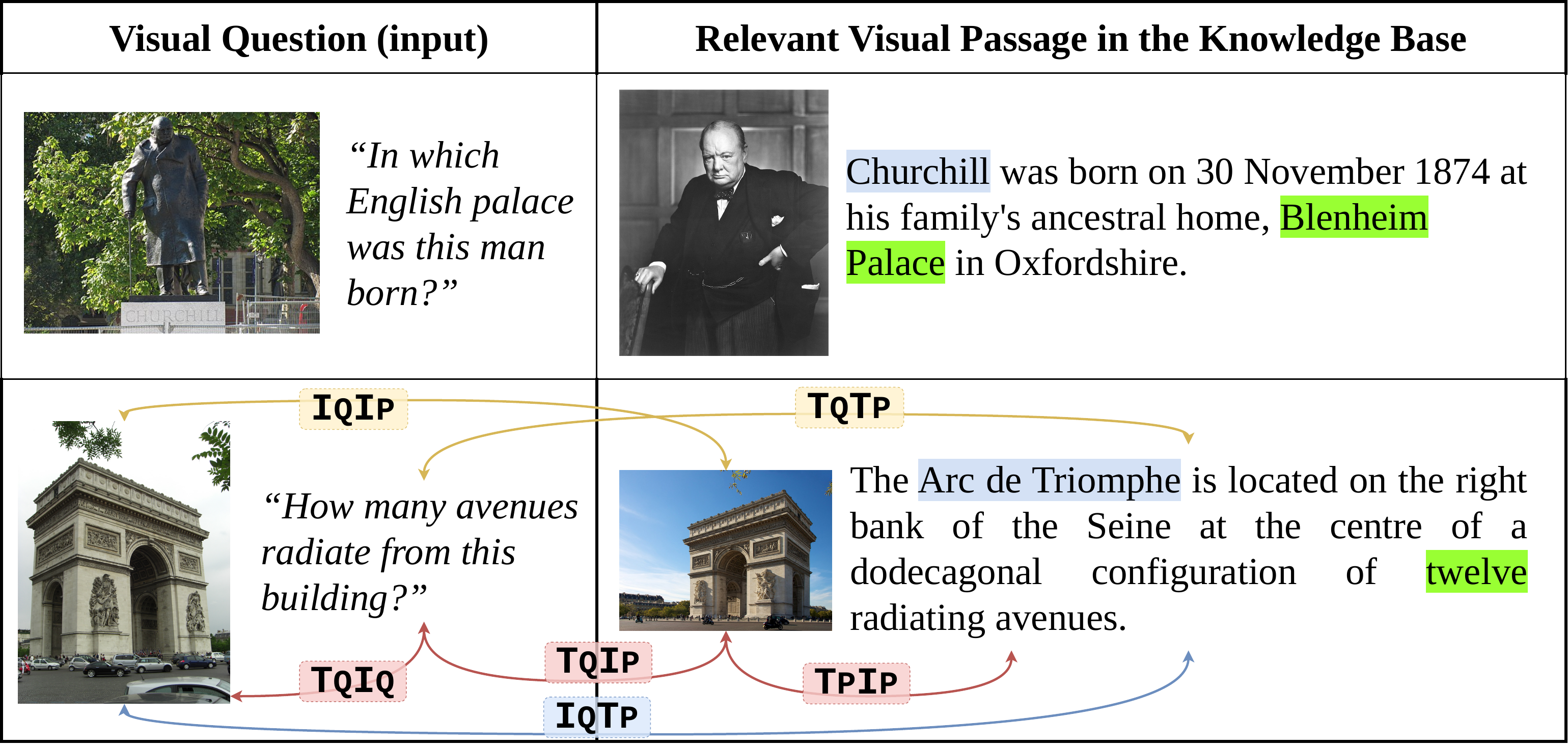} 
\caption{Two visual questions from the ViQuAE dataset along with relevant visual passages from its Knowledge Base.  The different types of mono- and cross-modal interactions studied are also shown for the second question. The acronyms of the interactions are composed of the letters \textsc{T} (Text), \textsc{I} (Image), \textsc{q} (question) and \textsc{p}~(passage).} 
\label{fig:viquae_eg_FR_cm_vs_mm} 
\end{figure}

The work we present in this article takes place in the context of Multimodal Information Retrieval, a field at the intersection between Information Retrieval (IR), Computer Vision, and Machine Learning. More precisely, we focus on Knowledge-based Visual Question Answering about named Entities (KVQAE), which has two specificities in regards to multimodal modeling \cite{bokhari_multimodal_2013,zhang2019information,baltrusaitis_multimodal_2019,guo_deep_2019}:
(i)~images represent named entities; (ii) multimodal interactions are complex and may be combined as both questions and retrieved passages are (text, image) pairs.
Indeed, KVQAE consists in answering questions about named entities grounded in a visual context \cite{shah2019kvqa,sigir2022}. We focus here on Entity Retrieval based on this visual context, similarly to Visual Named Entity Linking \cite{sun_visual_2022}.
Figure~\ref{fig:viquae_eg_FR_cm_vs_mm} shows two examples of visual questions along with corresponding relevant visual passages from the ViQuAE dataset \cite{sigir2022} and its multimodal Knowledge Base (KB), i.e. the set of multimedia documents in which the answers to the questions are searched.

The first example shows how heterogeneous depictions of named entities can be: Winston Churchill is depicted through a \textit{statue} in the visual question and by a \textit{standard photograph} in the KB. This heterogeneity makes mono-modal image retrieval difficult. On the other hand, cross-modal retrieval may bridge the semantic gap between the two representations by using a more abstract representation of the entity, e.g. its name, here \textit{Winston Churchill}. 

We formalize these different multimodal interactions in the framework exemplified in Figure~\ref{fig:viquae_eg_FR_cm_vs_mm}. The cross-modal interaction between the image of the question and the text of the passage is noted \IQTP{}, while the mono-modal interaction between the two images is noted \IQIP{}. This work is inspired by \cite{ecir2023} who studied early multimodal fusion methods, also modeling the \TQIQ{} (resp. \TPIP{}) interaction within the visual question (resp. passage), but found that \IQTP{} was the most important multimodal interaction. 

KVQAE differs from standard Visual Question Answering (VQA \cite{antol_vqa_2015}), which targets the content of the image (e.g., ‘‘\textit{What color is the car?}’’), and therefore does not require IR. Commonsense VQA~\cite{marino2019ok,schwenk_okvqa_2022} falls in between standard VQA and KVQAE but (i) focuses on Commonsense Knowledge; (ii) is limited to coarse-grained object categories, e.g., \textit{person} and \textit{building}, instead of \textit{Winston Churchill} and \textit{Arc de Triomphe}, which makes image retrieval straightforward using an object detector \cite{garderes_conceptbert_2020}.

KVQAE was introduced in \cite{shah2019kvqa} and received an increased interest recently, with a shift towards unstructured KBs \cite{sigir2022,ecir2023} and later Large Language Models (LLMs), which do not use an explicit KB but rather generate an answer from the knowledge implicitly stored in their parameters \cite{chen_can_2023,hu_avis_2023,Mensink_2023_ICCV,li_m3it_2023}. Given the results of \cite{chen_can_2023,Mensink_2023_ICCV} and the various caveats of LLMs for factual information generation (hallucinations, lack of generalization and updatability \cite{ji_survey_2023,zamani_retrieval_2022}), our work adopts a more classical Question Answering architecture, also exploited by \cite{sigir2022,ecir2023}, in which a first IR step is followed by an answer extraction step.

More precisely, we focus on Entity Retrieval and propose to use a multimodal dual encoder \cite{gan_visionlanguage_2022}, namely CLIP \cite{radford_learning_2021}, for both mono- and cross-modal retrieval, i.e. modeling \IQIP{} and \IQTP{}, respectively. 
Multimodal dual encoders like CLIP are used as foundation models for a set of diverse tasks such as multimodal analogy \cite{couairon_embedding_2022}, Visual Named Entity Linking  \cite{sun_visual_2022}, Cross-modal Question Answering \cite{liu_universal_2022}, and Commonsense VQA \cite{gui_kat_2022}. We show that both mono- and cross-modal retrieval are complementary and can be simply yet effectively combined. We provide empirical evidence through experiments on the ViQuAE, InfoSeek,
and Encyclopedic-VQA datasets, being as such the first comparative study of these recently introduced datasets. 
Furthermore, we study three different strategies to fine-tune such a model, which has been pre-trained in a cross-modal fashion, in this context: mono-modal, cross-modal, or joint training.

\section{Related Work} 
In this section, we present a review of datasets and methods for KVQAE.

\subsubsection{Datasets} 
KVQA was the first KVQAE dataset proposed in \cite{shah2019kvqa}. Despite its apparent large size, it has several limitations as pointed out by \cite{sigir2022}: (i) only one entity type is considered, namely \textit{person}; (ii) it is generated automatically, and thus, has a limited diversity of topics, lexicon, and syntax. Another key difference with the other datasets is that KVQA was designed for structured KBs, in particular Wikidata, from which it was generated, and not an unstructured KB like the following works. 
To address the limitations of KVQA, ViQuAE was introduced in \cite{sigir2022}. It has fewer visual questions but they are manually annotated and it covers a broad range of topics, lexicon, and syntax, as showed in Table~\ref{tab:data_stats}. Above all, ViQuAE comprises a large number of different entity types, including for example landmarks and organizations in addition to persons.
Recently, two other datasets were proposed, aiming at larger size than ViQuAE and with fewer textual bias: InfoSeek \cite{chen_can_2023} and Encyclopedic-VQA (EVQA \cite{Mensink_2023_ICCV}). InfoSeek is split into two subsets according to the annotation method: manual (ISM) or automatic (ISA). Unfortunately, since neither ISM nor the test set of ISA is available at the time of writing, we can evaluate our model only on the validation set of ISA. 
As its annotation is automatic, it shares part of the caveats of KVQA but covers more diverse entity types.
EVQA alleviates these by using more sophisticated question generation techniques than templates. However, it is sometimes biased towards text, with questions such as ``\textit{Which republic celebrated the vendémiaire in the month that the growing season ends for this tree?}'', a type of overspecified questions that were typically filtered by the manual annotation in ViQuAE \cite{sigir2022}. 
Some key features of these datasets are summarized in Table~\ref{tab:data_stats}. Question length is expressed in number of words provided by spaCy's English tokenizer. Answer prior is computed as the most likely answer in the training set, independently of the question. 
All datasets are limited to the English language. 

\begin{table}[t]
    \centering
    
    \caption{Key features of different KVQAE datasets: ViQuAE \cite{sigir2022}, InfoSeek \cite{chen_can_2023}, Encyclopedic-VQA (EVQA \cite{Mensink_2023_ICCV}), and KVQA \cite{shah2019kvqa}. InfoSeek is split into two subsets according to the annotation method: manual (ISM) or automatic (ISA). %
    *Computed on a subset of 500 questions by \cite{chen_can_2023}.
    }
    \label{tab:data_stats}
    \begin{tabular}{lrrrrr}
     
    \toprule
         & \textbf{ViQuAE} & \textbf{ISM} & \textbf{ISA} &
         \textbf{EVQA}&
         \textbf{KVQA}  \\ 
         \midrule
        \textbf{\# Visual questions} & 3,700 & 8,900&1,356,000& 
        1,036,000&
        183,000 \\ %
        \textbf{\# Unique questions (text-only)} & 3,562 & 2,022& 1,498& 175,000& 8,310 \\
        \textbf{\# Unique POS sequences} & 2,759 & 1,056& 267& 91,945& 376\\
        \textbf{\# Questions per image} & 1.1 & 1.0 & 1.4 & 2.0&7.4 \\ %
        \textbf{Vocabulary (\# words)} & 4,700 & 1,307&725 & 40,787 & 8,400\\
        \textbf{Average  question length (\# words)} & 12.4 & 7.8 & 8.9& 11.6  & 10.1 \\
        \textbf{Answer prior} & 0.3\%& -- & 0.6\% & 0.4\% & 15.9\% \\
        \textbf{Answer overlap} & 25.3\% & -- & 48.1\%&59.6\%& 89.4\%\\
        \textbf{Entity overlap} & 18.1\% & -- & 20.1\%&82.0\%& 40.6\% \\
        \textbf{\# Questions per entity} & 1.5 & 11.0 & 117.6 &62.5 & 9.7 \\ %
        \textbf{\# Entity types} & 980 & 527& 2,739& --&1 \\
        \textbf{Requires knowledge}* & 95.2\% & 95.6\% & --& -- & -- \\
        \bottomrule
    \end{tabular}
\end{table}

\subsubsection{Methods}
Because the KVQA dataset is limited to \textit{person}-named entities, it was addressed through face recognition in \cite{shah2019kvqa}:  a Wikidata subgraph is  constructed from the recognized entities and processed by a \textit{memory network} to output an answer \cite{weston2015memory}. A few other studies were carried out on KVQA but the comparison with the rest of the state of the art is made difficult as  their systems take the image \textit{caption} as input, making the image \textit{itself} redundant \cite{vickers_factuality_2021,garcia_improving_2021,heo_hypergraph_2022}.

Our work is closer to \cite{sigir2022}, which uses an unstructured KB, a collection of visual passages (as in Figure~\ref{fig:viquae_eg_FR_cm_vs_mm}). The authors tackle the task in two steps, where Reading Comprehension follows IR. Their retrieval is a combination of two mono-modal retrievals: textual with DPR \cite{karpukhin2020dense} and visual with a combination of CLIP, ArcFace \cite{deng_arcface_2019}, and a ResNet model trained on ImageNet \cite{he_deep_2016,deng_imagenet_2009}. We aim at simplifying this system by (i) removing the dependency on ArcFace and ImageNet, two supervised models that provide \textit{a priori} less generic representations than CLIP; (ii) taking full advantage of CLIP by combining mono-modal and cross-modal retrieval. After the IR step, answers are extracted using Multi-passage BERT  \cite{wang_multi-passage_2019}. This work was then extended in \cite{adjali2023icmr}, by combining the text retrieval of DPR with Wikidata embeddings, but in doing so, it sets aside multimodal interactions and the image of the visual question. 
For their part, \cite{ecir2023} have, like us, focused on IR. In order to model cross-modal interactions, they jointly represent text and image using a multimodal Transformer \cite{khan_transformers_2021,gan_visionlanguage_2022}. However, this model requires an expensive pre-training and the authors ultimately suggest that it mostly leverages  the  \IQTP{} interaction. Our conclusions converge because our model outperforms theirs --- without additional pre-training --- by explicitly modeling \IQTP{} via CLIP, as described in the next section. 

Very recently, following the overall trend in our domains, there has been a handful of works aiming to tackle KVQAE with (multimodal) LLMs, directly generating an answer from the visual question, without IR \cite{chen_can_2023,hu_avis_2023,Mensink_2023_ICCV,li_m3it_2023}. The same conclusions are reached in \cite{chen_can_2023} and \cite{Mensink_2023_ICCV}: multimodal LLMs suffer from the same caveats as text-only LLMs and underperform compared to retrieval-augmented models. As a consequence, a sophisticated planning method using a tool-augmented LLM as agent was proposed in \cite{hu_avis_2023}. However, \cite{hu_avis_2023} and \cite{Mensink_2023_ICCV} share the same experimental protocol problem: they query the whole Web for image or text retrieval through Google APIs, although the images of the visual questions are public and indexed by Google, which leads to overoptimistic and non-reproducible results. On the contrary, we follow the methodology of \cite{sigir2022,ecir2023,chen_can_2023}, using a controlled, publicly available KB. As for \cite{li_m3it_2023}, they tackle KVQAE with a multimodal LLM, which is only able to generate long explanatory answers. Therefore, \cite{li_m3it_2023} evaluate it on ViQuAE using ROUGE-L \cite{lin2004rouge}, after paraphrasing the ground-truth answers with ChatGPT. For that reason, their results are unfortunately not comparable with the rest of the state of the art. %

\section{Entity Retrieval from Visual Context} 
\subsection{Method}\label{sec:Methods}
Before being able to extract the answer to the question from a visual passage, or even retrieve such a passage, we focus here on Entity Retrieval, given the image of the question $\mathbf{i_q}$ and a collection of entities $(\mathbf{t_p},\mathbf{i_p})$, where $\mathbf{t_p}$ denotes the name of the entity and $\mathbf{i_p}$ its reference image. To do so, we define the following similarity function, which combines mono- and cross-modal similarities: \begin{equation} 
s(\mathbf{i_q}, \mathbf{t_p},\mathbf{i_p}) = \alpha_I s_I(\mathbf{i_q}, \mathbf{i_p}) + \alpha_C s_C(\mathbf{i_q}, \mathbf{t_p}) \label{eq:jcm_clip} 
\end{equation} 
where the parameters $\alpha_{\{I,C\}}$ weigh each similarity. We focus on CLIP, a multimodal dual encoder, to implement $s_I(\mathbf{i_q}, \mathbf{i_p})$ and $s_C(\mathbf{i_q}, \mathbf{t_p})$, which models the \IQIP{} and \IQTP{} interactions, respectively (see Figure~\ref{fig:viquae_eg_FR_cm_vs_mm}). The objective is thus to bring the image of the question closer to the image of this entity in the KB (\emph{mono-modal training}), or to its name (\emph{cross-modal training}), or both jointly. 

More formally, the objective underlying our IR model is to maximize $s(\mathbf{i_q}, \mathbf{t_p}, \mathbf{i_p})$ if the two images $\mathbf{i_q}$ and $\mathbf{i_p}^{(+)}$ depict the same entity, named with the textual form $\mathbf{t_p}^{(+)}$, and to minimize it otherwise. In such a contrastive approach, the other entities of a batch, for which the textual and visual representations are respectively noted $\mathbf{t_p}^{(j)}$ and $\mathbf{i_p}^{(j)}$, are used as negatives. To implement this approach, we jointly train $s_I(\mathbf{i_q}, \mathbf{i_p})$ and $s_C(\mathbf{i_q}, \mathbf{t_p})$ for each $\mathbf{i_q}$ image of the batch by minimizing the following objective, given the temperature $\tau$: 
\begin{equation}
-\log \frac{
    \exp{\left(s(\mathbf{i_q}, \mathbf{t_p}^{(+)},\mathbf{i_p}^{(+)})e^\tau\right)}}
    {\exp{\left(s(\mathbf{i_q}, \mathbf{t_p}^{(+)},\mathbf{i_p}^{(+)})e^\tau\right)} + \sum_{j} 
        \exp{\left(s(\mathbf{i_q}, \mathbf{t_p}^{(j)},\mathbf{i_p}^{(j)})e^\tau\right)}
    }
\label{eq:batch_contrastive}
\end{equation}
Since we implement $s_C(\mathbf{i_q}, \mathbf{t_p})$ with CLIP, we have: \begin{equation}
    s_C(\mathbf{i_q}, \mathbf{t_p}) = \cos\left(\clipv{}(\mathbf{i_q}), \clipt{}(\mathbf{t_p})\right)
\end{equation}
If $\alpha_I=0$ and $\alpha_C=1$ (cross-modal training only), the  objective is equivalent to the one used during the pre-training of CLIP, except that it is asymmetric (the softmax function expresses the probabilities according to $\mathbf{i_q}$ and not according to $\mathbf{t_p}$). Since $\mathbf{i_q}$, $\mathbf{t_p}$, and $\mathbf{i_p}$ are encoded independently, this objective leverages all the other images and texts of the batch in a highly efficient way (we only need a matrix product to compute the denominator of Equation~\ref{eq:batch_contrastive}). We implement $s_I(\mathbf{i_q}, \mathbf{i_p})$ in a similar way: $s_I(\mathbf{i_q} \mathbf{i_p}) = \cos\left(\clipv{}(\mathbf{i_q}), \clipv{}(\mathbf{i_p})\right)$. The same method could be applied to any multimodal dual encoder \cite{gan_visionlanguage_2022}. 

\subsection{Data} \label{ssec:Data}

As mentioned in the introduction, our evaluations are performed on the ViQuAE, ISA, and EVQA datasets. For ViQuAE and ISA, we use the KB proposed in \cite{sigir2022}, which consists of 1.5 million Wikipedia articles and images of corresponding Wikidata entities.
Unfortunately, the KB proposed by \cite{chen_can_2023} has yet to be made available; so our results on ISA will not be directly comparable to theirs. Indeed, 11.5\% of ISA entities are missing from our KB, which filters down the training set by 28\%. On the contrary, only a few entities from ViQuAE are missing from the KB. For EVQA, we use the corresponding KB of \cite{Mensink_2023_ICCV}, which consists of 2~million Wikipedia articles and corresponding images in WIT \cite{srinivasan_wit_2021}. 

ViQuAE contains \numprint{3700} visual questions about 2,400 different entities, randomly divided into equal-sized sets for training, validation, and testing, with no overlap between images. As a result, the overlap between entities in the training and test sets is quite small, only 18\%. Likewise, the entity overlap in ISA is of 20\%. Our models must therefore learn to generalize not only to new images but also to new entities.
On the contrary, the entity overlap of EVQA is of 82\%. 

\subsection{Hyperparameters} 
We use the \texttt{ViT-B/32} version of CLIP unless otherwise mentioned. 
To take full advantage of the entities associated with the other images in the batch $\mathbf{t_p}^{(j)}$ and $\mathbf{i_p}^{(j)}$, we use a batch of the largest possible size, here \numprint{3072}  $(\mathbf{i_q}, \mathbf{t_p}^{(+)},\mathbf{i_p}^{(+)})$ triples, i.e., more than the whole training set of ViQuAE. We use a single NVIDIA V100 GPU with 32~GB of RAM. The large batch size is partly enabled by gradient checkpointing. 

Because the training set of ViQuAE is so small, training is very cheap: our best model converges, i.e., starts to overfit, after 11 steps/epochs, in less than 15 minutes, which is negligible compared to the pre-training of 8,000 steps in three days of \cite{ecir2023} with the same hardware (\cite{ecir2023} reports a carbon footprint of 1.7 kgCO2e for 3 days of GPU power consumption). On the larger ISA and EVQA datasets, our models converge roughly after 500 steps in 5 hours.

We use a very small learning rate of $2 \times 10^{-6}$, increasing linearly for 4 steps and then decreasing for 50 steps on ViQuAE (or 1,000 on ISA and EVQA) if training is not interrupted before. We use the AdamW optimizer \cite{loshchilov-iclr-19}, with a weight decay of 0.1. For joint training, we initialize $\alpha_I = \alpha_C = \num{0.5}$ and assign them a learning rate of $\num{0.02}$, much larger than the rest of the model. Like \cite{radford_learning_2021}, the temperature $\tau$ remains trainable but, given the small learning rate, it remains close to its initial value, $\num{4.6}$.\footnote{We kept the formulation of \cite{radford_learning_2021} but the temperature is usually expressed as $\frac{1}{\tau'}$ and not $e^\tau$, which would be equivalent to $\tau'=\frac{1}{100}$ here.} These hyperparameters were set manually through experiments on the validation set of ViQuAE. 

Early stopping is done according to the \textit{in-batch} mean reciprocal rank on the validation set, i.e., by reranking the images or texts of the batch according to the similarity score $s$, to avoid computing the representations of the whole KB at each epoch.

Our implementation is based on Lightning,\footnote{\url{https://www.pytorchlightning.ai/}} PyTorch \cite{paszke_pytorch_2019}, and Transformers \cite{wolf_huggingfaces_2020} for training, and Datasets \cite{lhoest_datasets_2021}, Faiss \cite{johnson_billion-scale_2019}, and Ranx \cite{bassani_ranx_2022} for IR, based on the codebase of \cite{sigir2022}. Our code is freely available at \url{https://github.com/PaulLerner/ViQuAE} to ensure the reproducibility of our results.

\subsection{Results}\label{sec:Results}

We evaluate Entity Retrieval according to the relevance of the Wikipedia article associated with the target entity, which is determined automatically according to the presence of the answer after standard preprocessing (lowercasing, stripping articles, and punctuation). Additionally, because ISA contains a large portion of numerical answers, we follow the same soft matching method as \cite{chen_can_2023} for ISA (years can be off by one and there is a 10\% tolerance for measures and various numerical answers). We focus on the single-hop questions of EVQA, following \cite{Mensink_2023_ICCV}.
The metrics used are Precision at 1 (P@1) and Mean Reciprocal Rank (MRR).\footnote{The results are consistent with precision and recall at higher cutoffs, which we omit for the sake of space.}

\begin{figure}[t]
    \centering
    \includegraphics[width=.8\textwidth]{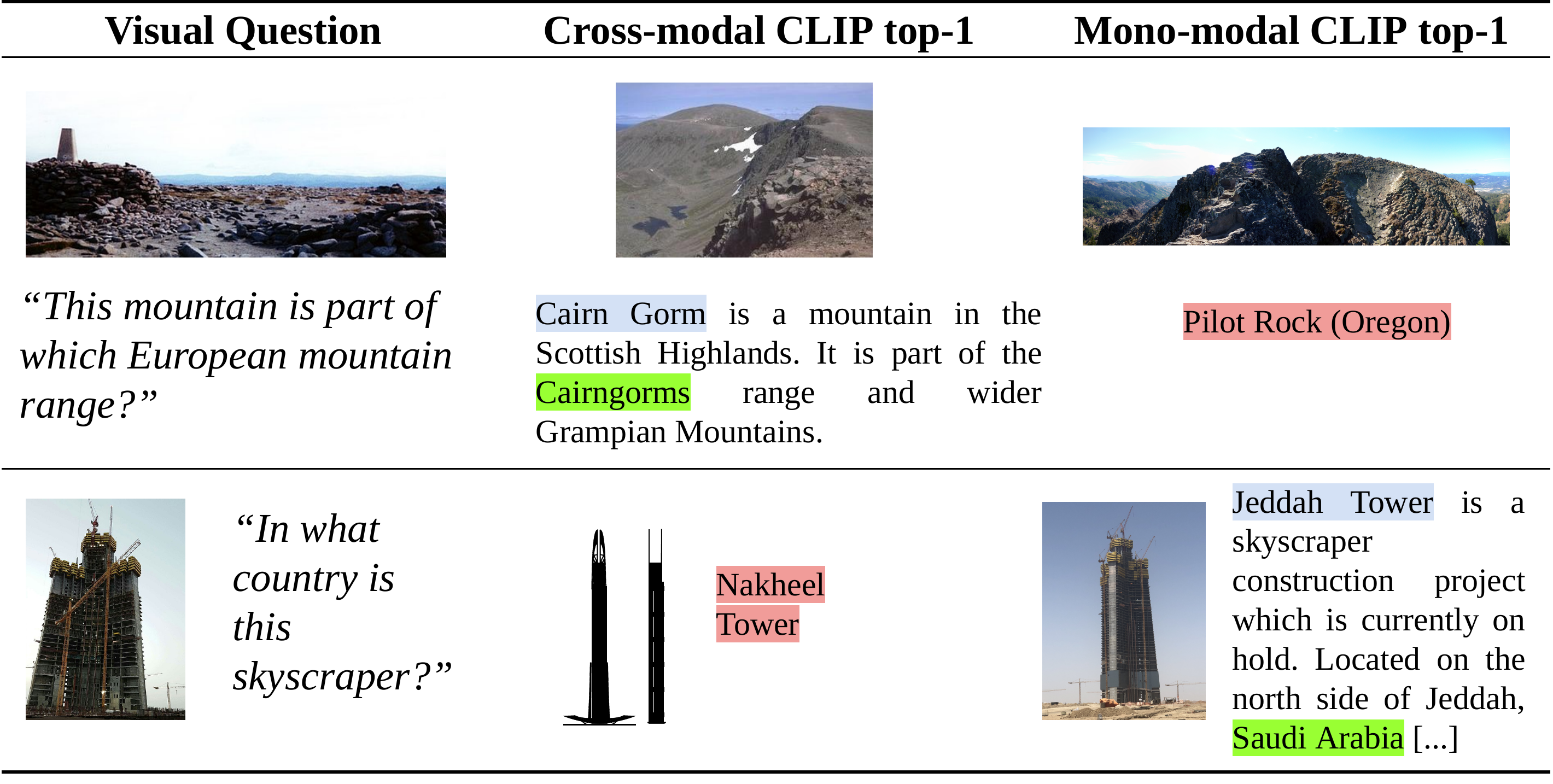}
    \caption{Strengths and weaknesses of mono- and cross-modal retrieval exemplified through CLIP results (not fine-tuned) on ViQuAE's validation set.}
    \label{fig:vitxvit_vs_vitxtitle_EN}
\end{figure}

\begin{table}[t] 
\caption{Entity Retrieval with a multimodal dual encoder, CLIP, on the validation subsets of ViQuAE, InfoSeek-Automatic (ISA), and  EVQA (single-hop). Mono- and cross-modal retrieval model the \IQIP{} and \IQTP{} interactions, respectively.  The best results are marked in bold for each type of retrieval. Hybrid retrieval of disjoint training combines \textit{mono-modal trained} mono-modal retrieval and \textit{cross-modal trained} cross-modal retrieval.}
\centering 

\begin{tabular}{c@{\extracolsep{.7em}}c@{\extracolsep{.7em}}cc@{\extracolsep{.7em}}cc@{\extracolsep{.7em}}cc} 
\toprule
\textbf{Retrieval} &\textbf{Training}& 
 \multicolumn{2}{c}{\textbf{ViQuAE}}  & \multicolumn{2}{c}{\textbf{ISA}}& \multicolumn{2}{c}{\textbf{EVQA}}\\
 \cmidrule{3-4}\cmidrule{5-6}\cmidrule{7-8}
& & \textbf{MRR}& \textbf{P@1}&  \textbf{MRR}& \textbf{P@1}&  \textbf{MRR}& \textbf{P@1}\\
\midrule
       \multirow{4}*{Mono-modal} & -- & 29.4 & 21.8 & 28.3 & 18.1 & 26.1 &             15.2 \\ 
        ~ & Mono-modal & 30.0 & 21.8 & \textbf{31.4} & \textbf{20.5}&    \textbf{32.6} &             \textbf{21.7} \\ 
        ~ & Cross-modal & 29.8 & 21.4 & 29.0 & 18.3   &29.7 &             18.9\\ 
        ~ & Joint & \textbf{30.4} & \textbf{22.0} & 30.5 & 19.9&   30.7 &             19.6 \\ 
        \midrule
        \multirow{4}*{Cross-modal} & -- & 32.7 & 23.1 & 32.8 & 22.4 &   20.9 &             12.2 \\ 
        ~ & Mono-modal & 31.6 & 21.9 & 33.0 & 22.0&      20.5 &             12.0 \\ 
        ~ & Cross-modal & \textbf{37.1} & \textbf{26.9} & \textbf{34.7} & \textbf{23.8}& \textbf{23.2} &             \textbf{13.8}  \\ 
        ~ & Joint & 30.8 & 21.3 & 31.1 & 20.3 &   22.4 &             12.9 \\ 
        \midrule
        \multirow{5}*{Hybrid} & -- & 39.6 & 30.6 & 36.2 & 25.8  &28.7 &             18.7\\ 
        ~ & Mono-modal & 40.1 & 31.8 & 38.2 & 27.4 &    33.8 &             22.9\\ 
        ~ & Cross-modal & \textbf{{44.1}} & \textbf{{34.9}} & 38.5 & 27.8&   33.8 &             23.3 \\ 
        ~ & Joint & 41.0 & 32.6 & 37.6 & 26.9 &    34.0 &             23.3  \\ 
        ~ & Disjoint & 43.7 & 34.5 & \textbf{40.0} & \textbf{29.6} &    \textbf{37.4} &             \textbf{27.8} \\ 
        \bottomrule
\end{tabular}%
\label{tab:image_ir_dev} 
\end{table}

We first explore in Table~\ref{tab:image_ir_dev} three training strategies and three ways of using a multimodal dual encoder through experiments conducted on the validation set. These three strategies can be defined from Equation~\ref{eq:jcm_clip}: 
\begin{itemize} 
\item Mono-modal (image-image) retrieval/training, i.e., $\alpha_I=1$, $\alpha_C=0$; 
\item Cross-modal (image-text) retrieval/training, i.e., $\alpha_I=0, \alpha_C=1$; 
\item Hybrid retrieval or joint training, i.e., $\alpha_I>0, \alpha_C>0$. 
\end{itemize} 
For hybrid retrieval, the weights $\alpha_{\{I,C\}}$ are set through a grid search over the validation set to maximize the mean reciprocal rank while constraining their sum to 1, so as to fairly compare joint training with mono- and cross-modal training. 
Note that the retrieval is independent from the training strategy, as shown in Table~\ref{tab:image_ir_dev}. Recall that CLIP's pre-training is only cross-modal~\cite{radford_learning_2021}, as most multimodal dual encoders \cite{gan_visionlanguage_2022}.

\subsubsection{Mono- or Cross-modal Retrieval?} Before comparing the different training methods, we can first notice that cross-modal IR  outperforms\footnote{Significantly according to Fisher’s randomization test \cite{fisher_design_1937,smucker_comparison_2007} with $p \leq 0.01$.}  mono-modal IR on both ViQuAE and ISA,\footnote{An exception is EVQA, for which mono-modal retrieval outperforms cross-modal retrieval. This is surprising as both EVQA and ISA stem from the iNaturalist \cite{van_horn_inaturalist_2018} and Google Landmarks \cite{weyand_google_2020} datasets. Further investigations are required.} %
especially  without fine-tuning (first lines of each block in Table~\ref{tab:image_ir_dev}), which may seem curious since proper nouns are not \textit{a priori} very meaningful. Therefore, it is surprising that CLIP generalizes\footnote{Unless its pre-training dataset contains enough entities from ViQuAE and ISA so that it circumvents generalization. We develop this discussion in Section~\ref{sec:Conclusion}.} to new entity names. Nevertheless, some names carry meaning. For example, a name can indicate the gender of a person or suggest their nationality.\footnote{An interactive visualization is provided at \url{https://paullerner.github.io/ViQuAE/\#text-embedding-cross-modal}.} Moreover, we are working here with titles of Wikipedia articles, which are also likely to contain the nature of the entity (e.g., the profession of a person or the type of a monument). These features can thus be mapped to visual attributes.

Foremost, we mainly attribute the success of cross-modal IR to its adequacy with the pre-training of CLIP: the representation space of CLIP is organized to bring together similar texts and images, which the mono-modal proximity of images is only an indirect consequence of. We show examples of successes and failures in Figure~\ref{fig:vitxvit_vs_vitxtitle_EN}. In line with the results of \cite{ecir2023}, we observe that mono-modal retrieval may be more sensitive to superficial image details (color vs. black-and-white photography, subject pose\ldots). Here, the two photographs at the top of two mountains, showing the horizon, are judged to be similar even though they are different mountains. In contrast, the mono-modal retrieval is more effective in the second example, where the two photographs of the Jeddah Tower are taken from similar vantage points. These qualitative results support our hypothesis that cross-modal retrieval might help addressing the heterogeneity of visual representations of named entities.

\subsubsection{Why choose?} We show that mono- and cross-modal retrievals are complementary: their results can be simply combined at the score level (as in Equation~\ref{eq:jcm_clip}). 
Thus, without fine-tuning (first lines of each block in Table~\ref{tab:image_ir_dev}), fusing the two retrievals brings a relative improvement of 32\% in P@1 for ViQuAE (and 15\% for ISA, 23\% for EVQA) compared to the best single retrieval (significant with Fisher's $p \leq \num{0.01}$). It would be interesting to study whether these results generalize to other tasks. For example, this method could benefit Content-based Image Retrieval in a Web browsing context. 
Overall, hybrid retrieval gives the best performance, on all three datasets.

\subsubsection{How to fine-tune multimodal dual encoders?}
We see that fine-tuning with a given strategy (e.g. mono-modal) always enhances the performance of retrieval with the same strategy. However, it also sometimes decreases retrieval with another strategy (e.g. cross-modal). Therefore, we find it best to combine models trained disjointly: a \textit{mono-modal trained} mono-modal retrieval and a \textit{cross-modal trained} cross-modal retrieval.

\section{Retrieving Passages and Extracting Answers}
\subsection{Methods}
While we have focused on Entity Retrieval through cross-modal retrieval, we are ultimately interested in answering questions about these entities. To do so, we follow the same framework as \cite{sigir2022}, where Entity Retrieval results are mapped to the corresponding passages to enable fusion with a text passage retrieval method, such as DPR.  

\begin{figure}[t]
    \centering
    \includegraphics[width=.4\textwidth,trim={7.9cm 1.5cm 3.5cm 2cm},clip]{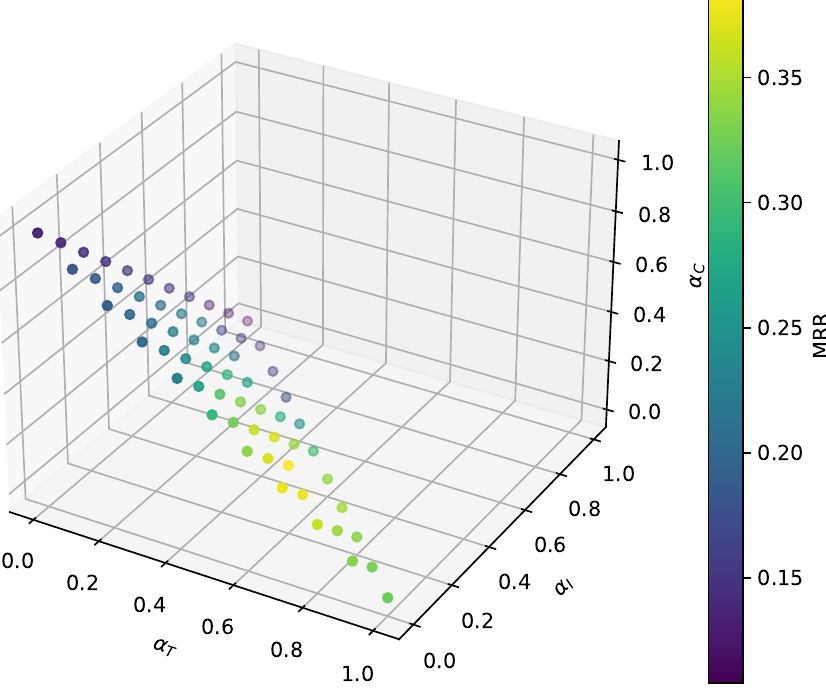}
    \caption{Passage-level MRR on the validation set of ViQuAE depending on the $\alpha_{\{T,I,C\}}$ hyperparameters.}
    \label{fig:mrr_wrt_alpha}
\end{figure}

This implies redefining $s$ as follows: 
\begin{equation}
    s(\mathbf{t_q},\mathbf{i_q}, \mathbf{t_p},\mathbf{i_p}) = \alpha_T s_T(\mathbf{t_q}, \mathbf{t_p}) + \alpha_I s_I(\mathbf{i_q}, \mathbf{i_p}) + \alpha_C s_C(\mathbf{i_q}, \mathbf{t_p})
    \label{eq:jcm}
\end{equation}
where $s_T(\mathbf{t_q}, \mathbf{t_p})$ models the \TQTP{} interaction between the text of the question and of the passage and is implemented with DPR. 
We note this model $\mathrm{DPR}_{V+T}$ as it combines DPR, $\clipv{}$, and $\clipt{}$, or $\mathrm{DPR}_{\mathbf{V}+\mathbf{T}}$ (in bold font) when CLIP is fine-tuned.\footnote{DPR is always fine-tuned as described in the next section.} The weights $\alpha_{\{T,I,C\}}$ are set through a grid search on the validation set like in the previous section (see Figure~\ref{fig:mrr_wrt_alpha} for an illustration of the impact of these hyperparameters on MRR).
DPR is a dual encoder model that combines two BERT encoders, one for the question and one for the passage \cite{karpukhin2020dense}. 

Answers are then extracted from these passages using  Multi-passage BERT \cite{wang_multi-passage_2019}, which also models the \TQTP{} interaction. 

\subsection{Data and implementation}
The 1.5 (resp. 2) million articles of the KB of ViQuAE \cite{sigir2022} (resp. EVQA \cite{Mensink_2023_ICCV}) are divided into 12 (resp. 27) million 100-word passages, while preserving sentence boundaries, as in \cite{sigir2022}. 

Both DPR and Multi-passage BERT are pre-trained on TriviaQA, filtered out of all questions used in \cite{sigir2022} to generate ViQuAE,\footnote{\url{https://huggingface.co/datasets/PaulLerner/triviaqa_for_viquae}} before being fine-tuned on the downstream KVQAE dataset, following \cite{sigir2022}.
Both models are built upon the uncased version of BERT-base \cite{devlin2018bert}. We refer the reader to \cite{sigir2022} for further implementation details.

\subsection{Baselines}\label{ssec:Baselines}

We compare our approach to the $\mathrm{DPR}_{V+R+A}$ model of \cite{sigir2022}, which combines DPR, $\clipv{}$, ArcFace, and an ImageNet-trained ResNet model.  The results of the four models are combined in the same way as in Equation~\ref{eq:jcm}, where DPR implements $s_T(\mathbf{t_q}, \mathbf{t_p})$, $\clipv{}$, ArcFace, and ImageNet compose $s_I(\mathbf{i_q}, \mathbf{i_p})$, and there is no cross-modal similarity, i.e., $s_C(\mathbf{i_q}, \mathbf{t_p})=0$. 

We also compare our methods to the $\mathrm{ECA}_{V}$ and $\mathrm{ILF}_{V}$ models of \cite{ecir2023}. ECA (Early Cross-Attention)  early-fuses modalities through an attention mechanism. The similarity is computed as $s(\mathbf{t_q},\mathbf{i_q}, \mathbf{t_p},\mathbf{i_p}) = \mathrm{ECA}(\mathbf{t_q},\mathbf{i_q}) \cdot \mathrm{ECA}(\mathbf{t_p},\mathbf{i_p})$ and thus combines all the multimodal interactions shown in Figure~\ref{fig:viquae_eg_FR_cm_vs_mm}. ILF (Intermediate Linear Fusion) fuses modalities with a simple linear projection and thus has, like our method, neither \TQIQ{} nor \TPIP{} interactions since the similarity can be reduced to: 
\begin{equation}
    s(\mathbf{t_q},\mathbf{i_q}, \mathbf{t_p},\mathbf{i_p}) = s_T(\mathbf{t_q}, \mathbf{t_p}) + s_{C'}(\mathbf{t_q}, \mathbf{i_p}) + s_I(\mathbf{i_q}, \mathbf{i_p}) +  s_C(\mathbf{i_q}, \mathbf{t_p})
\end{equation}
Note that \cite{sigir2022,ecir2023} use $\clipv$ with the ResNet architecture while we use ViT \cite{dosovitskiy_image_2020} in most of our experiments (but compare the two in the next section and find no significant difference). 
    
Moving away from the Retrieval+Extraction framework of \cite{sigir2022}, we compare our results to \cite{chen_can_2023,Mensink_2023_ICCV}, who both use the PaLM LLM \cite{chowdhery2023palm}, either as is or augmented with the image caption and in-context learning examples (denoted PromptCap \cite{hu2023promptcap}). \cite{chen_can_2023} also experiment with FiD \cite{izacard_leveraging_2021}, augmented with CLIP retrieval results.

\subsection{Results}\label{sec:rc_results}

\begin{table}[t]

\caption{Reading Comprehension results on the test set of ViQuAE, the validation set of ISA, and the test single-hop questions of EVQA. As in \cite{karpukhin2020dense}, the reader takes as input the top-24 of different IR systems listed in the ``Method'' column (except for the methods of \cite{chen_can_2023,Mensink_2023_ICCV}). The results of \cite{chen_can_2023}, in gray, are provided as reference but use a different, yet unavailable, smaller KB, which perfectly covers ISA.
*CLIP is based on ViT’s architecture instead of ResNet. \textdagger{}Our re-implementation of the reader, which fixes the loss function.}
\label{tab:rc_viquae_results_yes_triviaqa}

\centering
\resizebox{\columnwidth}{!}{
\begin{tabular}{lrcc@{\extracolsep{.7em}}c@{\extracolsep{.7em}}cc}
\toprule
\textbf{Method}& \textbf{\# Param. (M)}& \multicolumn{2}{c}{\textbf{ViQuAE}} & \textbf{ISA} & \multicolumn{2}{c}{\textbf{EVQA}}\\

        \cmidrule{3-4} \cmidrule{5-5} \cmidrule{6-7}
&&\textbf{EM} &    \textbf{F1} & \textbf{Soft Match} & \textbf{BEM} & \textbf{F1} \\ 
\midrule
PaLM few-shot (text-only) \cite{chen_can_2023} & 540,000& 31.5 & -- & \hphantom{0}4.8& --& --\\
CLIP + FiD \cite{chen_can_2023} & 1,170 %
& -- & -- &\textcolor{gray}{20.9}& --& -- \\
PaLM zero-shot (text-only) \cite{Mensink_2023_ICCV} & 540,000& --& --& --&19.7&--\\
PromptCap + PaLM \cite{Mensink_2023_ICCV} 
& 540,870 %
 & --& --& --&29.7&--\\
\midrule

 DPR (text-only) \cite{sigir2022} & 330 %
 &16.9  &    20.1 &--& --& -- \\

$\mathrm{DPR}_{V}$ \cite{ecir2023} & 432%
& 19.0 &    22.3  &--& --& -- \\
$\mathrm{DPR}_{V}$* (baseline) &481& 19.7 	& 23.3 &--& --& -- \\
$\mathrm{DPR}_{V}$*\textdagger{} (baseline) &481 & 26.4 & 29.1&\hphantom{0}7.7&  27.4&25.4 \\
 $\mathrm{DPR}_{V+R+A}$ \cite{sigir2022} & 500%
 &22.1  &  25.4 &--& --& -- \\
$\mathrm{ECA}_{V}$ \cite{ecir2023} & 432& 20.6  &    24.4&--& --& --  \\
$\mathrm{ILF}_{V}$ \cite{ecir2023}& 432 &21.3  & 25.4  &--& --& --\\
$\mathrm{DPR}_{\mathbf{V}+\mathbf{T}}$* (this work) &481 & {24.7}  &{28.7} &-- & --& --\\
$\mathrm{DPR}_{\mathbf{V}+\mathbf{T}}$*\textdagger{} (this work)&481 & \textbf{30.9} & \textbf{34.3} & \textbf{12.4} & \textbf{29.1}&  \textbf{26.6}  \\
\midrule
Oracle retrieval + FiD \cite{chen_can_2023} & Oracle + 770& -- & -- & \textcolor{gray}{52.5}& --& --\\

Oracle retrieval\textdagger{} (this work)&Oracle + 110& 68.3 & 72.7 & 46.8 &65.3&59.7 \\
\bottomrule
\end{tabular}
}
\end{table}

\subsubsection{Metrics} Extracted answers  are evaluated using Exact Match (EM) and token-level F1 score on ViQuAE following \cite{sigir2022}, using the soft matching score defined by \cite{chen_can_2023} on ISA (see Section~\ref{sec:Results}), and using both F1 and BEM \cite{bulian-etal-2022-tomayto} on EVQA. The results for these three benchmarks are reported in Table~\ref{tab:rc_viquae_results_yes_triviaqa}.

\subsubsection{Hybrid retrieval effectiveness}
When comparing the $\mathrm{DPR}_{V}$ and $\mathrm{DPR}_{\mathbf{V}+\mathbf{T}}$ models, we see that the effectiveness of combining mono- and cross-modal retrieval observed earlier indeed translates to more accurate answers, on all three datasets. Therefore, our model also outperforms the previously proposed models of \cite{sigir2022,ecir2023} on ViQuAE, while being conceptually simpler and computationally cheaper (emitting hundred times less CO2 than \cite{ecir2023}). 
Furthermore, we found a bug in the implementation of the reader's loss provided by \cite{sigir2022}. Fixing it consistently improved results, for both $\mathrm{DPR}_{V}$ and $\mathrm{DPR}_{\mathbf{V}+\mathbf{T}}$. Our model is also competitive with the method of \cite{Mensink_2023_ICCV} on EVQA, while using \numprint{1000} times less parameters.\footnote{We focus on the single-hop subset of EVQA following \cite{Mensink_2023_ICCV}. On the two-hop questions, the model using $\mathrm{DPR}_{\mathbf{V}+\mathbf{T}}$ achieves 31.1 BEM and 25.6 F1, and 9.8 BEM/3.8 F1 on the multi-answer questions.}

\subsubsection{Knowledge base incompleteness}
The results of \cite{chen_can_2023} are provided as reference but are hardly comparable to the others. Apart from PaLM being three order of magnitude greater than the other models and partly trained on ViQuAE's test set,\footnote{According to \cite{chowdhery2023palm}, around 20\% of TriviaQA is contained in PaLM’s pre-training dataset. ViQuAE was derived from TriviaQA \cite{sigir2022}.} they use a different KB. This KB, yet unavailable, is fifteen times smaller than ours, so contains less distractors, and covers 100\% of the entities and questions of ISA. In contrast, our KB lacks 11.5\% of ISA entities and is not guaranteed to contain the answers for the 88.5\% remaining, because of differences between the Wikipedia versions.

\subsubsection{Oracle retrieval}
We conduct additional experiments in an ``oracle retrieval'' setting, where the reader only takes relevant passages as input, similarly to the oracle experiments of \cite{sigir2022,chen_can_2023}. In agreement with their results, we find a large gap between our best retrieval model and the oracle, showing that IR is still the main bottleneck of KVQAE. Compared to the FiD model of \cite{chen_can_2023}, with 770M parameters, we approach its performance, although Multi-passage BERT is seven times smaller and our KB does not fully cover ISA.

\section{Conclusion}\label{sec:Conclusion}
This paper studies cross-modal retrieval and its combination with mono-modal retrieval for Knowledge-based Visual Question Answering about named Entities (KVQAE). Retrieval is carried out with a multimodal dual encoder, namely CLIP. Our results demonstrate the superiority of cross-modal retrieval over mono-modal retrieval, but also the complementarity of the two, which can be easily combined. 

We argue that cross-modal retrieval may help addressing the heterogeneity of visual representations of named entities, consistently with prior work.
It would be interesting to study whether these results generalize to other tasks. For example, this method could benefit Content-based Image Retrieval, in a Web browsing context.

Although it was the abundance of cross-modal data that enabled CLIP’s training in the first place, which would have been difficult with a mono-modal annotation, this limits our results because it is difficult to control such a large amount of data and thus to estimate CLIP's generalization capabilities. We hypothesize that mono-modal retrieval is better suited to generalize to new entities. 

We show that the effectiveness of cross-modal retrieval leads to more accurate answers, on all three studied datasets. Therefore, our method outperforms our baseline (mono-modal retrieval) but also the methods of \cite{sigir2022,ecir2023}, while being conceptually simpler and computationally cheaper. Furthermore, it is competitive with billion-scale parameters models on ISA and EVQA. 
As such, this is the first comparative study of the recently introduced ViQuAE, ISA, and EVQA datasets. We find that ISA is more challenging as it is less biased towards text, but advocate for further studies on all three datasets --- which all have their pros and cons --- with diverse methods. 

Consistently with \cite{sigir2022,chen_can_2023}, we find a large gap between our best retrieval model and oracle retrieval, showing that entity retrieval is the main bottleneck of KVQAE. For future work, we plan to combine our unstructured KB with a structured one, such as Wikidata, to enable the modeling of links between the entities \cite{xie_image-embodied_2017,pezeshkpour_embedding_2018,wilcke_end--end_2020,alberts_visualsem_2020}, which would further address the heterogeneity of their visual representations. A more IR perspective on the matter could cast KVQAE as a query expansion problem, with an initial ambiguous textual query which would benefit from pseudo-relevant feedback \cite{Xu1996Query}.

\bibliographystyle{splncs04}
\bibliography{Bibliography,biblio-local}
\end{document}